%% file: main.tex
\documentclass[letterpaper, 10 pt, conference]{ieeeconf}
\IEEEoverridecommandlockouts                              
                                                          
\usepackage{times}
\usepackage{epsfig}
\usepackage{graphicx}

\usepackage[T1]{fontenc}
\usepackage[utf8]{inputenc} 

\usepackage{booktabs}
\usepackage{array,multirow,graphicx} 
\usepackage{bbding} 
\usepackage{amsmath} 
\usepackage{amssymb}  
\usepackage{floatrow}
\usepackage{algorithm,algpseudocode}
\usepackage{caption}
\usepackage{subcaption}
\usepackage{float}

\usepackage{color}

\usepackage{flushend}
\usepackage{placeins}

\usepackage{siunitx} 
\providecommand{\etal}{\textit{et~al.}}

\usepackage[dvipsnames]{xcolor}
\usepackage[colorlinks,bookmarksopen,bookmarksnumbered,citecolor=Blue,linkcolor=Blue,urlcolor=red]{hyperref}
\usepackage[capitalize,nameinlink,noabbrev]{cleveref} 

\title{A water-obstacle separation and refinement network for unmanned surface vehicles}

\author{Borja Bovcon$^1$ and Matej Kristan$^1$%
\thanks{*This work was supported in part by the Slovenian research agency (ARRS) programmes P2-0214 and P2-0095, and the Slovenian research gency (ARRS) research project J2-8175.}%
\thanks{$^{1}$Borja Bovcon and Matej Kristan are with University of Ljubljana, Faculty of Computer and Information Science, Slovenia\newline
        {\tt\small \{borja.bovcon, matej.kristan\}@fri.uni-lj.si}}%
}
\begin{document}

\maketitle
\thispagestyle{empty}
\pagestyle{empty}

\begin{abstract} 

Obstacle detection by semantic segmentation shows a great promise for autonomous navigation in unmanned surface vehicles (USV). However, existing methods suffer from poor estimation of the water edge in presence of visual ambiguities, poor detection of small obstacles and high false-positive rate on water reflections and wakes. We propose a new deep encoder-decoder architecture, a water-obstacle separation and refinement network (WaSR), to address these issues. Detection and water edge accuracy are improved by a novel decoder that gradually fuses inertial information from IMU with the visual features from the encoder. In addition, a novel loss function is designed to increase the separation between water and obstacle features early on in the network. Subsequently, the capacity of the remaining layers in the decoder is better utilised, leading to a significant reduction in false positives and increased true positives. Experimental results show that WaSR outperforms the current state-of-the-art by a large margin, yielding a 14\% increase in F-measure over the second-best method.

\end{abstract}



\section{Introduction}
\label{sec:intro}
Recent developments in field robotics inaugurated a new class of small-sized unmanned surface vehicles (USVs). These vessels are ideal for operation in coastal waters and narrow marinas due to their portability, and can be used for automated inspection of hazardous and difficult to reach areas. Uninterrupted and safe navigation requires a high level of autonomy. One of the main challenges in autonomous navigation is timely detection and avoidance of near-by obstacles. Various sensors have been considered for this task (e.g., RADAR~\cite{Onunka2010}, LIDAR~\cite{Jimenez2009}, SONAR~\cite{Heidarsson2011}), among which cameras have shown a great potential as affordable, lightweight and powerful obstacle detection devices~\cite{KristanCYB2015,cane2016saliency,bb_iros_2018,prasad2018overview,muhovivc2019obstacle}.
 
Traditional maritime  camera-based obstacle detection methods rely on background subtraction~\cite{prasad2018overview}, but these are not appropriate for USVs due inherent scene dynamics, making the system non-robust and prone to false-positive detections. Stereo-based reconstruction methods~\cite{muhovivc2019obstacle,wang2013stereovision} address the dynamic environment, but require sufficiently textured scene and obstacles that significantly stick out of the water. Calm, poorly textured water and flat floating objects thus lead to detection failure. Stereo baselines have to be kept small to maintain the USV stability, which also reduces the detection range. Semantic segmentation methods based on fitting a structured models to the image~\cite{KristanCYB2015,bb_ras_2018,bb_iros_2018} have achieved excellent results and are currently the state-of-the-art on this domain. But these approaches rely on simple features which fail to fully capture the scene appearance diversity. Segmentation quality is thus degraded particularly in the presence of visual ambiguities and reflections~\cite{bb_ras_2018}.
  
\begin{figure}
	\centering
      \includegraphics[width=0.95\textwidth]{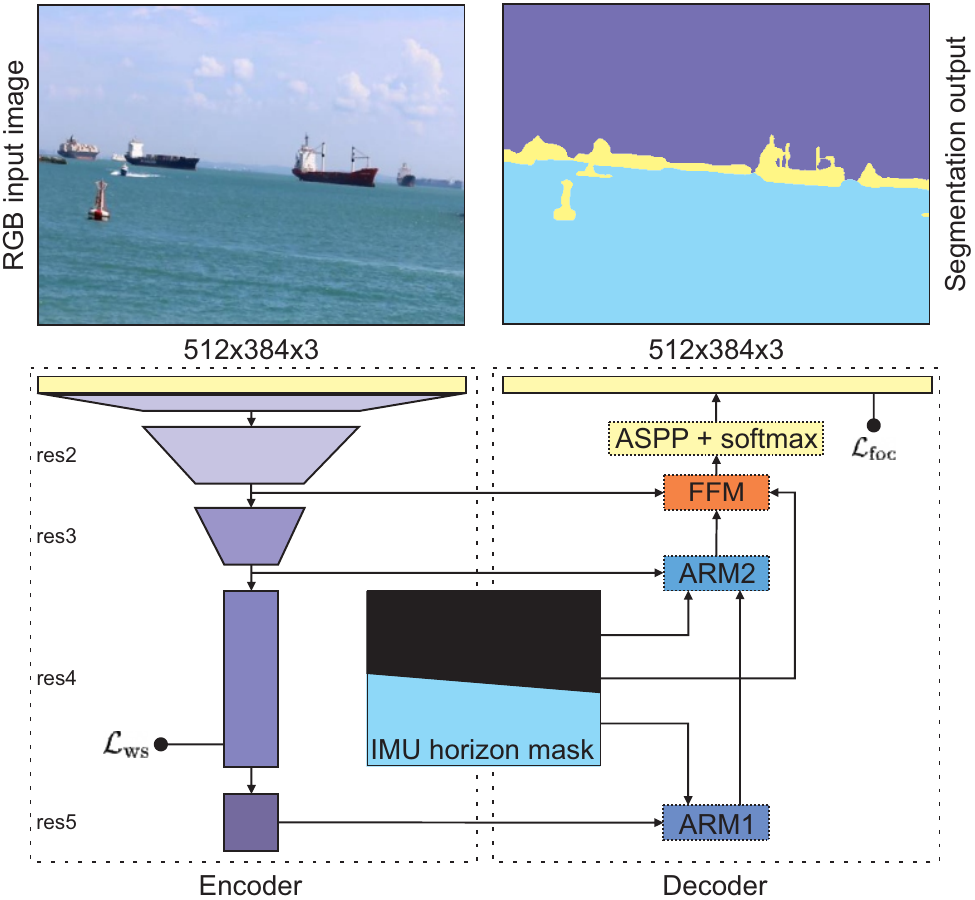}
      \caption{Architecture of the proposed WaSR network. Encoder generates rich deep features, which are gradually fused in the decoder with a horizon mask computed from an IMU readout to boost detection and water edge estimation. A water-obstacle separation loss $\mathcal{L}_\mathrm{WS}$ computed at the end of encoder drives learning of discriminative features, further reducing false positives and increasing true positives.  
      }
      \label{fig:intro_image}
\end{figure}  
  
Richer features can be learned by deep convolutional neural nets, and indeed developments in autonomous ground vehicles (AGVs)~\cite{chen2016deeplab,zhao2017pyramid,yu2018bisenet,chen2018encoder}, have demonstrated that these methods achieve remarkable semantic segmentation results.
But due to many differences between the AGV and USV domain, these networks cannot be readily applied to USVs. Most obvious difference is that the navigable surface in a maritime domain (water) is non-flat, dynamic, varies significantly in appearance and is greatly affected by the weather conditions.    
   
A recent study~\cite{bb_iros_2019} analyzed the performance of state-of-the-art AGV segmentation networks on a maritime domain. The study has shown that these networks, when trained on a large maritime dataset, outperform, or perform on par, with the model-based segmentation approaches~\cite{KristanCYB2015,bb_iros_2019}, but several issues remain. The large water appearance variability causes poor estimation of the water edge, and produces many false positives. Worse yet, the networks were often missing small obstacles, which leads to dangerous false negative detections.

Following the findings from~\cite{bb_ras_2018,bb_iros_2019} we propose a novel water-obstacle separation and refinement network (WaSR) designed as an encoder-decoder architecture (\cref{fig:intro_image}). A deep encoder is used to extract rich features from the input image, while a shallow decoder is used to gradually refine the segmentation. Our first contribution is  fusion of the external inertial sensory data from IMU with the visual information in the decoder, leading to a more accurate water edge segmentation and overall improved detection. 
Our second contribution is a new water-obstacle separation loss that aids learning of features that compactly encode a range of the water appearances, while enforcing a separation from the features corresponding to the obstacles. The loss is applied in a late stage of the encoder to foster learning discriminative features and simplify learning of the subsequent classifiers in the decoder.
WaSR shows impressive results on the currently most challenging USV dataset and sets a new state-of-the-art in USV obstacle detection.

\section{Related work}
\label{sec:relatedWork}
Cameras, combined with computer vision algorithms, have proven as a powerful yet affordable obstacle detection devices~\cite{larson2007advances,KristanCYB2015,cane2016saliency,bb_iros_2018,prasad2018overview,muhovivc2019obstacle}. Numerous image-processing methods for obstacle detection have been proposed. In-depth experimental evaluation of background subtraction methods~\cite{prasad2018overview}, has shown that misleading dynamics of water cause a great amount of false-positive detections. Stereo reconstruction methods~\cite{huntsberger2011stereo,wang2013stereovision,muhovivc2019obstacle}, are only capable of detecting obstacles well above the water surface. Poorly textured and partially submerged objects are likely to be undetected, while the detection of distant obstacles largely depends on the stereo baseline. On the other hand, semantic segmentation methods, based on fitting a structured models to the image~\cite{KristanCYB2015,bb_ras_2018,bb_iros_2018}, have achieved promising results and are capable of detecting obstacles protruding through the water surface, as well as the floating and distant ones. However, these approaches rely on simple features which fail to correctly address the diversity of a marine scene, thus leading to a poor segmentation in the presence of visual artefacts on water (wakes, sea foam, glitter, reflections, etc.).

Deep convolutional neural networks enable extraction of richer features, mandatory for accurate segmentation in the presence of visual ambiguities. Their training procedure requires a huge amount of carefully annotated data. Therefore, a large variety of urban datasets~\cite{cityscapes_dataset,kitti_dataset,yu2018bdd100k} have greatly contributed to a rapid development of deep neural nets~\cite{chen2016deeplab,yu2018bisenet,chen2018encoder,Jeong2018_psp} for AGVs which achieve astonishing segmentation results. However, due to many differences between the AGV and USV domain, these networks cannot be readily applied to USVs. For instance, the navigable surface in a maritime domain (water) is non-flat, extremely dynamic and varies significantly in its appearance. Moreover, turbulent waters cause USVs to rotate around roll-axis, while ground vehicles do not experience this phenomenon. 

Nonetheless,~\cite{lee2018image} and~\cite{yang2019surface} proposed using Faster R-CNN~\cite{ren2015faster} in their approach to detect and classify different types of ships.
However, Faster R-CNN cannot detect arbitrary obstacles without providing additional training data. Alternatively,~\cite{zhan2019autonomous} suggested an online segmentation approach for water component extraction. 
The segmentation accuracy continuously improves as online training progresses. However, their method requires a long and non-autonomous ``calibration'' procedure to start producing satisfactory results.

Recently, two separate studies~\cite{cane2018evaluating,bb_iros_2019} have evaluated the performance of commonly used deep segmentation networks from AGV domain on the task of obstacle detection in maritime surveillance. Cane~\etal~\cite{cane2018evaluating} used a filtered ADE20k~\cite{zhao2017pyramid} dataset for training and several maritime datasets (MODD~\cite{KristanCYB2015}, IPATCH~\cite{ipatch_dataset}, SEAGULL~\cite{seagull_dataset} and SMD~\cite{smd_prasad2017video}) for evaluation. On the other hand, Bovcon~\etal~\cite{bb_iros_2019} trained the nets on their proposed pixel-wise annotated maritime dataset (MaSTr1325) and perfomed the evaluation on MODD2~\cite{bb_ras_2018}. In both studies, evaluated methods have shown consistent drawbacks in water component segmentation and mis-classification of smaller obstacles.

\section{Semantic segmentation network WaSR}\label{sec:usvnet}
The architecture of WaSR is described in~\cref{sec:architecture}, a new water-obstacle separation loss is described in~\cref{sec:loss_function}, and~ 
\cref{sec:postprocessing} describes conversion of the segmentation result into obstacle detection output.

%

\subsection{Architecture overview}\label{sec:architecture}
The proposed WaSR (\cref{fig:intro_image}) architecture consists of a \textit{contracting path} (encoder) and an \textit{expansive path} (decoder). The purpose of the encoder is construction of deep rich features, while the primary task of the decoder is fusion of inertial and visual information, increasing the spatial resolution and producing the segmentation output.
 
Following the recent analysis~\cite{bb_iros_2019} of deep networks on a maritime segmentation task, we base our encoder on the low-to-mid level backbone parts of DeepLab2~\cite{chen2016deeplab}, i.e., a ResNet-101~\cite{he2016deep} backbone with atrous convolutions. In particular, the model is composed of four residual convolutional blocks (denoted as \textit{res2, res3, res4} and \textit{res5}) combined with max-pooling layers (see~\cref{fig:intro_image}). Hybrid atrous convolutions are added to the last two blocks for increasing the receptive field and encoding a local context information into deep features.
 
One of primary tasks of the decoder is fusion of visual and inertial information. 
We introduce the inertial information by constructing an IMU feature channel that encodes location of horizon at a pixel level. In particular, camera-IMU projection~\cite{bb_ras_2018} is used to estimate the horizon line and a binary mask with all pixels below the horizon set to one is constructed (\cref{fig:intro_image}). This IMU mask serves a prior probability of water location and for improving the estimated location of the water edge in the output segmentation. 

The IMU mask is treated as an externally generated feature channel, which is fused with the encoder features at multiple levels of the decoder. However, the values in the IMU channel and the encoder features are at different scales. To avoid having to manually adjust the fusion weights, we apply an approach called Attention Refinement Modules (ARM) proposed by~\cite{yu2018bisenet} to learn an optimal fusion strategy. 

\begin{figure}
	\centering
      \includegraphics[width=0.9\textwidth]{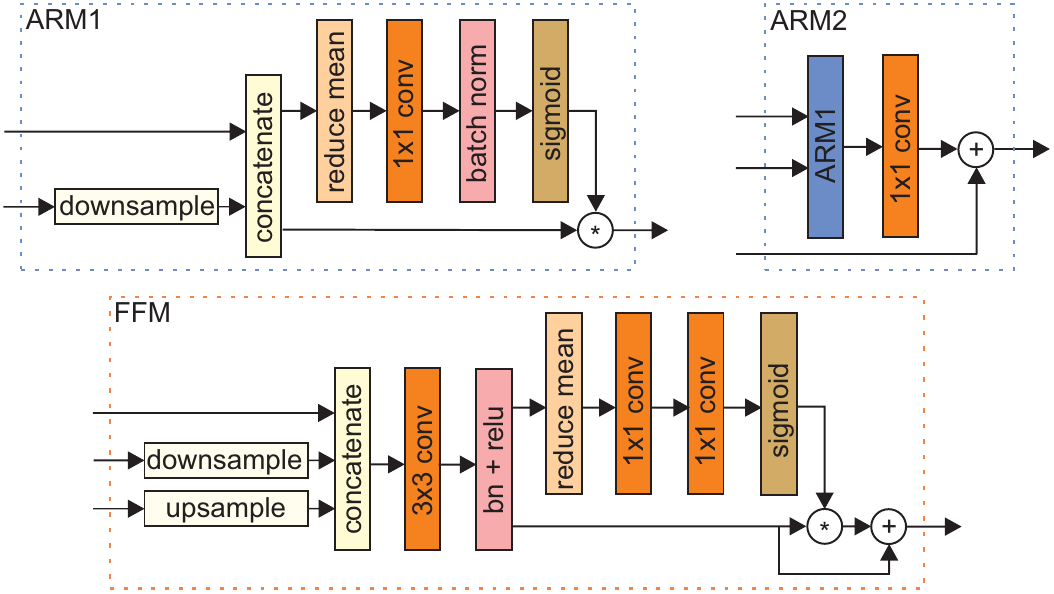}
      \caption{Attention refinement modules ARM1, ARM2 and feature fusion module FFM adjust the scale of heterogeneous input feature channels and gradually fuse inertial and visual information in the WaSR decoder.}
      \label{fig:modules_visualisation}
\end{figure}

The decoder starts with the ARM1 block (\cref{fig:modules_visualisation}), which differs from ARM~\cite{yu2018bisenet} in the way the input is pre-processed. The IMU mask is resized and concatenated with the encoder output features. The remaining steps follow~\cite{yu2018bisenet}: global average pooling followed by depth reduction and normalization is used to learn channel weights, which are subsequently used to re-weight the concatenated feature channels. The resulting features are further fused with res3 output features and the IMU mask using another ARM block called ARM2 (\cref{fig:modules_visualisation}). ARM2 first applies an ARM1 block to fuse the IMU mask and the features from lower part of the decoder. This is followed by a set of $1 \times 1$ convolutions to double the number of feature channels, which are per-channel summed with the \textit{res3} features from the encoder.

Yu et al.~\cite{yu2018bisenet} have argued the benefits of using a learnable fusion technique called Feature Fusion Module (FFM) for fusing low-level and high-level features in CNNs. In contrast to ARM, this module can implement fusion pathways of higher complexity. Our decoder thus up-samples the ARM2 output features and concatenates them with the \textit{res2} features and IMU mask. The depth of the resulting feature channels is halved by $3 \times 3$ convolution block and normalized by a batch-normalization block. A weight vector is then computed similarly to ARM1 and used to re-weight the features, leading to feature selection and fusion. 

Our recent study~\cite{bb_iros_2019} has shown that Atrous Spatial Pyramid Pooling~\cite{chen2016deeplab} (ASPP) leads to significant improvements in segmentation of small structures, yet entails only a small computational overhead. Thus the ASPP block, followed by a softmax, is added as the final block of our decoder. The resolution of the output features is quarter of the input resolution, forming a truncated U-shape net with skip connections. A smaller, non-symmetrical decoder contributes to the speed due to a lower amount of up-sampling procedures and convolutions. Finally, the decoder output is up-sampled by a factor of four to match the input resolution.

\subsection{Enforcing water-obstacle features separation}
\label{sec:loss_function}
%

Care has to be taken when designing a loss function for maritime environment. 
While some obstacles may be large, the majority of pixels in a typical marine scene belong either to water or sky. This leads to a class imbalance, which overwhelms the classical cross entropy loss. 
Furthermore, segmentation difficulty vastly ranges between different water regions. For example, it may be easy to classify regions of mildly rippled blue water, but it is much more difficult to classify glitter and mirrored reflections of objects in the water as the water component. Therefore, to adjust the focus of the network to challenging regions during training, we employ a focal loss~\cite{lin2017focal}, $\mathcal{L}_{\mathrm{foc}}$, adapted for segmentation. A classical L$_2$ loss, $\mathcal{L}_{L_2}$, is added for weight regularisation~\cite{krogh1992simple}.


Our recent study~\cite{bb_iros_2019} has shown that water appearances like glitter and object mirroring pose a significant challenge to water segmentation networks. While mistaking water for sky does not pose a threat, mistaking obstacles for water and vice versa does lead to a potential USV collision or frequent false alarms, rendering the network useless for practical navigation. To avoid this, the network should ideally learn encoding in early layers such that it produces very similar features for a variety of water appearances and very different features for obstacles. This makes the subsequent learning of the classifier in the higher layers of the network easier.

We propose enforcing early feature separation by designing a novel loss. Let $\{ x_j^c \}_{j \in O}$ and $\{ x_j^c \}_{j \in W}$ be features in channel $c$ belonging to pixels in the water region $W$ and the obstacle regions $O$, respectively. Since we would like to enforce clustering of water features, we can approximate their distribution by a Gaussian with per-channel means $\{ \mu^c \}_{c \in N_c}$ and variances $\{ \sigma^{c2} \}_{c \in N_c}$, where $N_c$ is the number of channels, and we assume channel independence for computational tractability. Similarity of all other pixels corresponding to obstacles can be measured as a joint probability under this Gaussian, i.e.,
\begin{equation}\label{eq:probwaterseparation}
    p( \{ x_j \}_{j \in W}) \propto \prod_{\substack{j\in W\\c=1:N_c}} exp(-0.5 (x_j^c - \mu^c)^2/\sigma^{c2}).
\end{equation}
We would like to enforce learning of features that minimize this probability. By expanding the equation for water per-channel standard deviations, taking the log of (\ref{eq:probwaterseparation}), flipping the sign and inverting, we arrive at the following equivalent obstacle-water separation loss
\begin{equation}\label{eq:water_separation_loss}
    \mathcal{L}_{\mathrm{ws}}=  \frac{N_O}{N_C N_W} \sum_c^{N_c} \frac{ \sum_{i \in W} (x_i^c - \mu^c)^2 } { \sum_{j \in O} (x_j^c - \mu^c)^2  },
\end{equation}
where the $N_O$ and $N_C$ are added as normalisation constants making the scale independent of the number of channels and obstacle pixels in individual frames. The final loss is a weighted summation of individual losses
\begin{equation}
    \mathcal{L} = \mathcal{L}_{\mathrm{foc}} + \lambda_1 \mathcal{L}_{\mathrm{ws}} + \lambda_2 \mathcal{L}_{\mathrm{L_2}}, 
\end{equation}
where $\lambda_1$ and $\lambda_2$ are the weights.

\subsection{Segmentation post-processing}
\label{sec:postprocessing}
\begin{figure}
	\centering
      \includegraphics[width=1\textwidth]{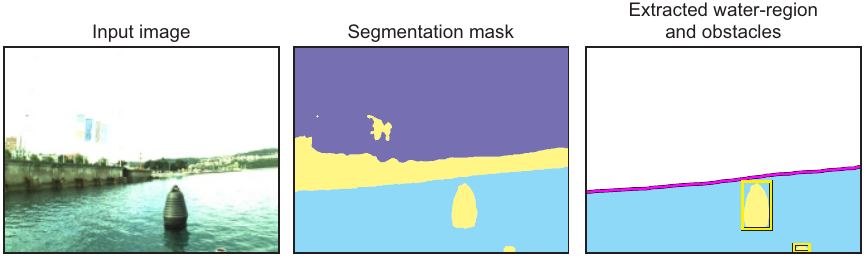}
      \caption{Raw image captured by the USV (left), WaSR segmentation output (middle) and post-processed segmentation output (right). Water, sky and obstacles are depicted with cyan, deep blue and yellow colour respectively. Extracted water-edge and obstacles are denoted with a pink line and yellow bounding boxes, respectively.}
      \label{fig:segmentation_postprocessing}
\end{figure}

The model proposed in~\cref{sec:architecture} outputs a segmentation mask where each pixel belongs to exactly one semantic component (water, sky or environment). Pixels marked with water label are used to construct the water-region mask as described in~\cite{bb_ras_2018}. 
%
%
The largest connected component in the water-region mask represents the navigable surface of the USV, and its upper edge corresponds to the edge of the water. The list of potential obstacles is obtained by extracting blobs of pixels marked with environment label within the water-region. The post-processing procedure and its results are illustrated in~\cref{fig:segmentation_postprocessing}.

\section{Experimental evaluation} \label{sec:experimental_evaluation}
The dataset and the evaluation protocol are described in~\Cref{sec:datasets}, the implementation details are given in~\cref{sec:implementation_details} and comparison to state-of-the-art and ablation study are given in~\cref{sec:results_comparison} and~\cref{sec:ablation_study}, respectively.

\subsection{Performance evaluation protocol and the dataset} \label{sec:datasets}

\begin{figure}
	\centering
      \includegraphics[width=1\textwidth]{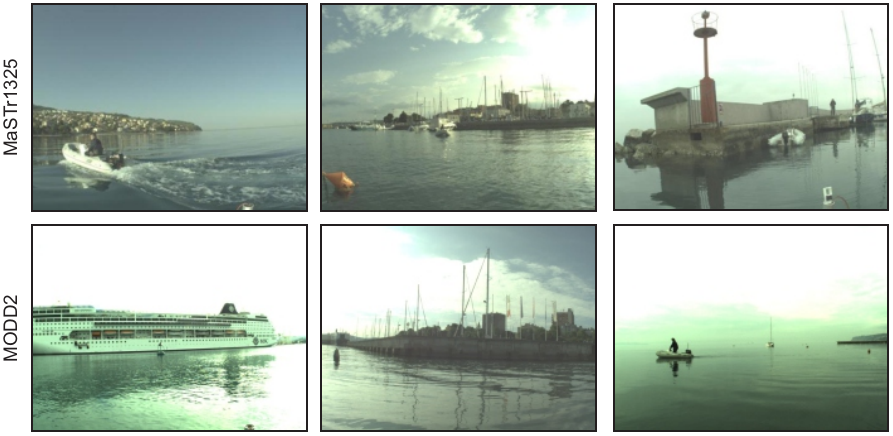}
      \caption{MaSTr1325~\cite{bb_iros_2019} (top) and Modd2~\cite{bb_ras_2018} (bottom) datasets exhibit a large scene and appearance variability.}
      \label{fig:dataset_images}
\end{figure}

We follow the recent protocol for evaluation of segmentation-based obstacle detectors in marine environment~\cite{bb_iros_2019}. The network is trained on the MaSTr1325 dataset~\cite{bb_iros_2019}, which is currently the largest annotated maritime segmentation dataset. The dataset was captured in a coastal sea area with a real USV  during a period of 24 months and consists of $1325$ high-resolution images ($1278 \times 958$ pixels) of various representative marine environment scenes (see \cref{fig:dataset_images} top row). Each image is per-pixel manually segmented by human annotators into three semantic components: sea, sky and obstacles. The edges of the semantically different components are labelled with the ``unknown'' category in order to address the annotation uncertainty and to allow automatic exclusion of these pixels from learning. Each image is equipped by a read-out from an IMU sensor on-board the USV.

Performance is evaluated on the Modd2 dataset~\cite{bb_ras_2018}, which is currently the most challenging public USV dataset due to a large variety of scenarios (object mirroring, glitter, and various weather conditions) present. Examples of images from this dataset are shown in the bottom of~\cref{fig:dataset_images}.
The dataset consists 28 stereo sequences, time-synchronised with measurements of the on-board IMU. Following the guidelines from~\cite{bb_iros_2019}, the left-camera image is used for evaluation. Obstacles and the water edge are manually annotated with bounding boxes and a polygon, respectively. 
 
As in MaSTr1325~\cite{bb_iros_2019}, we use the standard performance evaluation measures from~\cite{KristanCYB2015}. The accuracy of water-edge estimation is reported by mean-squared error computed over all sequences, while the accuracy of detected obstacles is measured by the number of true positives (TP), false positives (FP), false negatives (FN) and by the overall F-measure, i.e., a harmonic mean of precision and recall.

\subsection{Implementation details}
\label{sec:implementation_details}
Fast and accurate detection is crucial for autonomous systems. To gain speed, all input images were scaled to the resolution $512 \times 384$ pixels by bilinear interpolation. This resolution retains all hazardous obstacles  visible. Detections, as well as ground truth obstacles, with surface area of less than $5 \times 5$ pixels were ignored, since they do not pose a threat at the given resolution.

Dataset augmentation is used to increase generalisation capability of the trained networks. We applied vertical mirroring and central rotations of $\pm\left\lbrace 5, 15 \right\rbrace$ degrees on whole training images, while elastic deformation was applied solely on the water component of training images. Following~\cite{bb_iros_2019}, we also applied colour-transfer augmentation, resulting in total of $54325$ training images.

All networks were trained using a RMSProp optimizer with a momentum $0.9$, initial learning rate $10^{-4}$ and standard polynomial reduction decay of $0.9$. The weights of ResNet-101 backbone were pre-trained on ImageNet~\cite{deng2009imagenet}, while the remaining additional trainable parameters of our model (e.g., those from adding IMU channel and those in FFM, ARM and ASPP) were randomly initialised using
Xavier~\cite{glorot2010understanding}. The networks were fine-tuned on augmented training set for five epochs.


WaSR was implemented in Tensorflow\footnote{The reference implementation will be made publicly available on our project page.} and all experiments were run on a desktop computer with Intel Core i7-7700 3.6GHz CPU and nVidia GTX1080 Ti GPU with 11GB GRAM.

\subsection{Comparison with the state-of-the-art}
\label{sec:results_comparison}
%

WaSR from \cref{sec:usvnet} was compared to five recent state-of-the-art networks: PSPNet~\cite{zhao2017pyramid}, SegNet~\cite{badrinarayanan2017segnet} and BiSeNet~\cite{yu2018bisenet} were selected since they obtain state-of-the-art performance on segmentation tasks for autonomous cars, DeepLab3+~\cite{liu2018deep} (denoted as DL3+) was selected as state-of-the-art general-purpose segmentation network and a DeepLab variant called DeepLab2$_\mathrm{NOCRF}$~\cite{bb_iros_2019} (denoted as DL2$_\mathrm{NOCRF}$) was chosen since it achieved the best performance on a maritime segmentation problem~\cite{bb_iros_2019} among several networks. 
The results are summarised in~\cref{tab:results_detection_MODD2}.

\begin{figure*}
  \centering
  \includegraphics[width=1\textwidth]{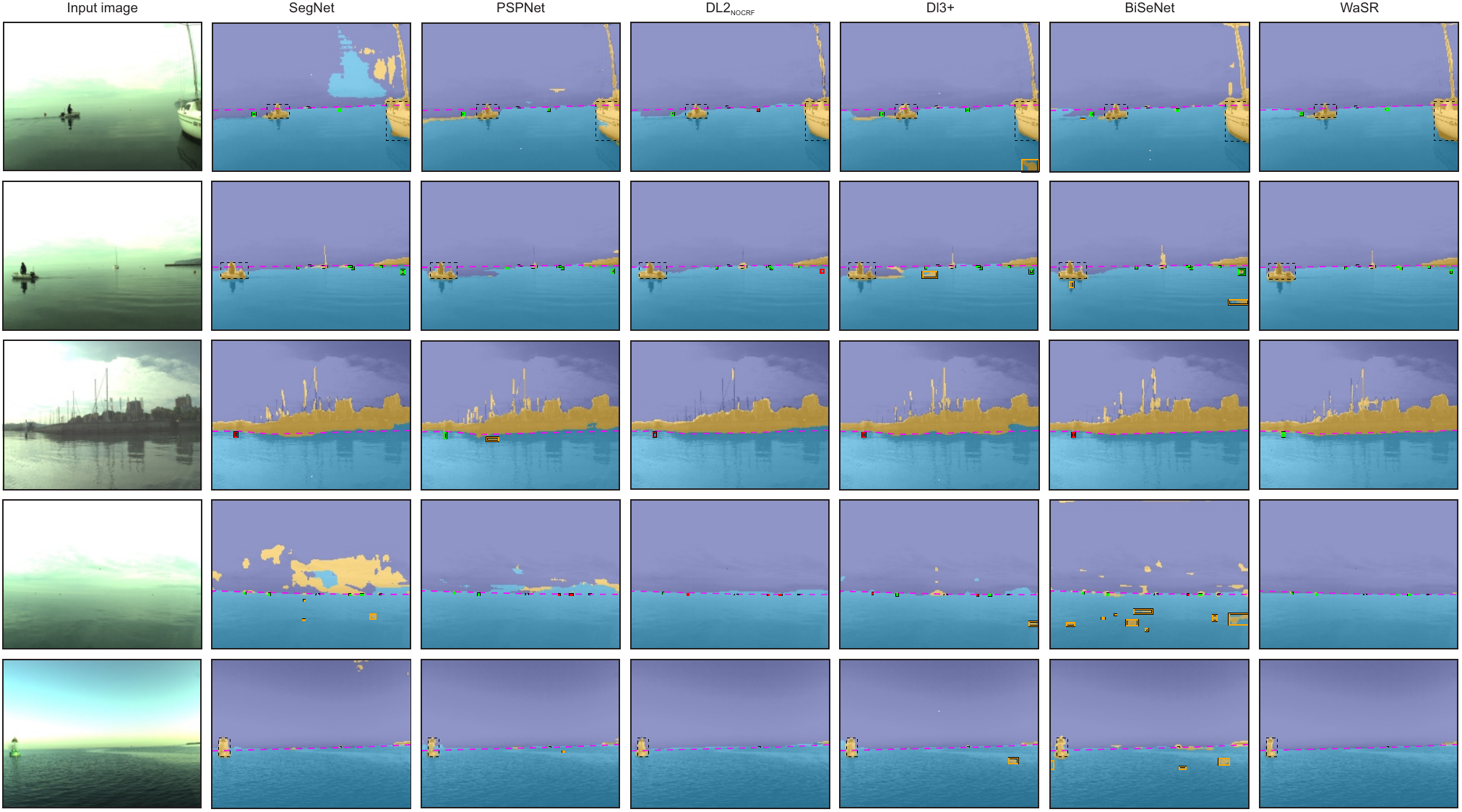}
  \caption{Qualitative comparison of segmentation quality. The sky, obstacles and water components are denoted with deep-blue, yellow and cyan colour, respectively. Correctly detected obstacles are marked with green bounding box, false positive detections with orange bounding box and undetected obstacles with red bounding box.}
  \label{fig:qualitative_comparison}
\end{figure*}

On the task of water-edge estimation, the proposed WaSR outperforms all other networks by a large margin. The second best is BiSeNet, lagging behind by three pixels worse accuracy, followed by DL2$_\mathrm{NOCRF}$, SegNet, PSPNet and DL3+. Visual inspection shows  that other networks struggle with accurately estimating the water edge in presence of haze on the horizon, while WaSR does neither overshoot nor undershoot its location. Some examples are shown in~\cref{fig:qualitative_comparison}. WaSR also shows impressive robustness to severe environmental mirroring in the water and estimates the water edge accurately even under these conditions (\cref{fig:qualitative_comparison} third row), while operating in real-time at approximately 10 frames-per-second. 

WaSR detects the highest number of true positives, followed by PSPNet, SegNet, BiSeNet, DL3+ and DL2$_\mathrm{NOCRF}$. Qualitative comparison (\cref{fig:qualitative_comparison} second, third and fourth row) shows that WaSR detects smaller obstacles more accurately than the other networks. While most of the other networks produce false positives on glitter, reflections and wakes, WaSR is largely robust to these and does not mistake them for obstacles (\cref{fig:qualitative_comparison} fourth and fifth row). 
A closer observation of first two rows in \cref{fig:qualitative_comparison} shows that the other networks perform poorly in presence of distinct wakes caused by boats. This results either in deteriorated water-edge estimation or false detections on the wake edges. Several networks experience noisy false detections across the image when the USV faces a hazy open-sea (\cref{fig:qualitative_comparison} fourth row), while WaSR remains unaffected. In fact, WaSR obtains the second-lowest false positive rate, tightly following DL2$_\mathrm{NOCRF}$, however, this is because DL2$_\mathrm{NOCRF}$ is prone to poor detection of isolated obstacles, leading to a high false negative rate and relatively low true positive rate.


\begin{table}
    \input{MODD2_detections}
    \caption{Results on Modd2~\cite{bb_ras_2018} report water-edge estimation error $\mu_{\mathrm{edg}}$ in pixels, the number of true positives (TP), false positives (FP), false negatives (FN) and the F-measure.}
    \label{tab:results_detection_MODD2}
\end{table}

\subsection{Ablation study}
\label{sec:ablation_study}

\begin{table}
    \input{MODD2_ablation}

    \caption{Ablation study results on Modd2~\cite{bb_ras_2018}, determining the importance of the IMU information and water-obstacle separation loss in the proposed architecture. We report the water-edge estimation error $\mu_{\mathrm{edg}}$, measured in pixels, the number of true positive (TP), false positive (FP), false negative (FN) detections and the F-measure.}
    \label{tab:results_ablation_MODD2}
\end{table}


The two major novelties in the WaSR architecture are (i) the object-water separation loss (\ref{eq:water_separation_loss}), and (ii) fusion of the external IMU sensor with the image data (\cref{sec:usvnet}). To evaluate the contribution of each, two variants of the WaSR were created and evaluated by the procedure from~\cref{sec:results_comparison}. The first variant was WaSR with the water-object separation loss removed (WaSR$_{\mathrm{NOWS}}$) and the second variant was WaSR with the IMU fusion removed (WaSR$_{\mathrm{NOIMU}}$). 
Results in~\cref{tab:results_ablation_MODD2} indicate that both, the separation loss and the IMU fusion importantly improve the performance.

\begin{figure}[ht!]
	\centering
      \includegraphics[width=1\textwidth]{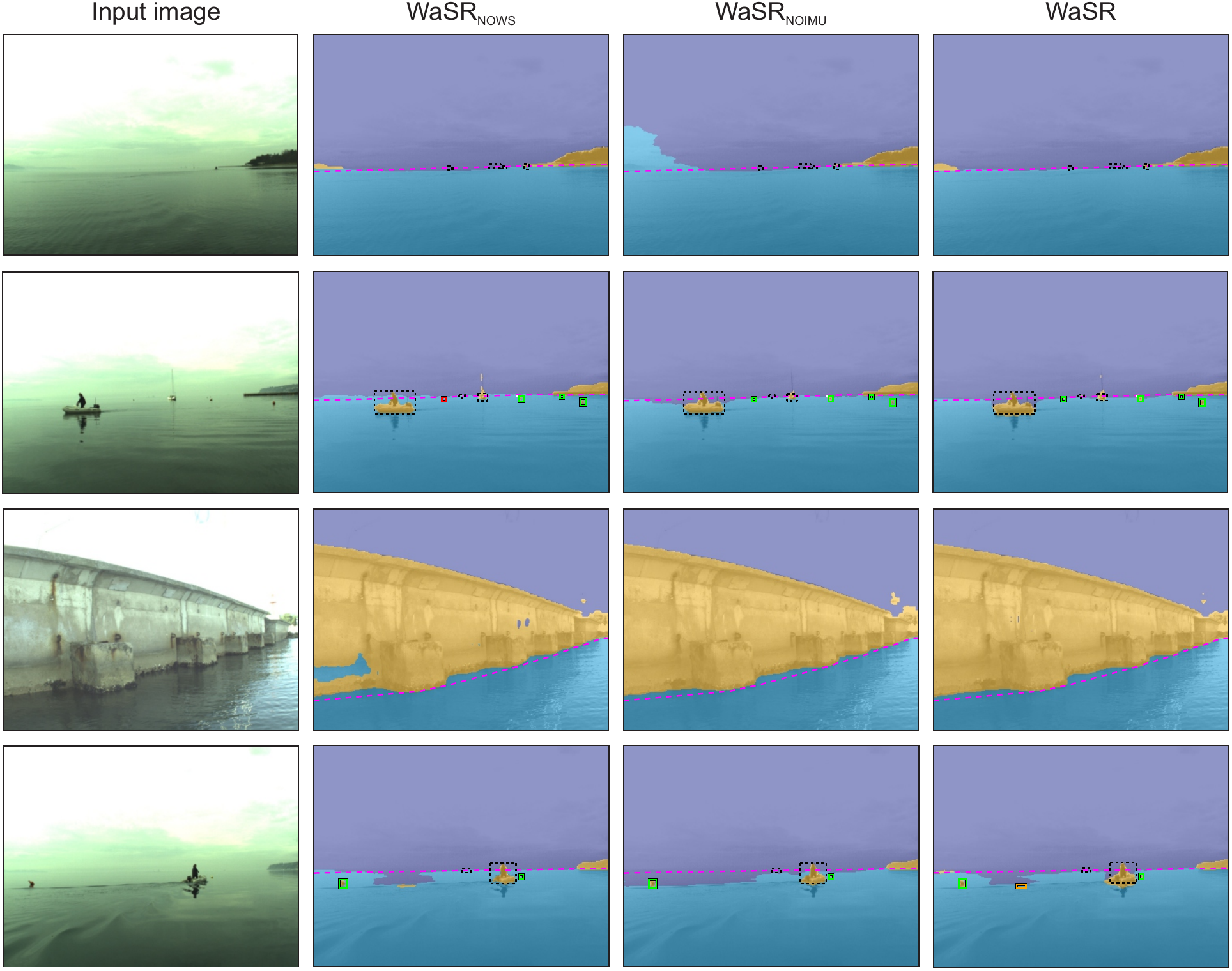}
      \caption{Qualitative analysis of the effects of using the water-obstacle separation loss and IMU fusion. The sky, obstacles and water are denoted by deep-blue, yellow and cyan, respectively. Detected obstacles are denoted by green (true positive), orange (false positive) and red (false negative).
      }
      \label{fig:ablation_study}
\end{figure}

A detailed inspection of~\cref{tab:results_ablation_MODD2} shows that the water-obstacle separation loss significantly improves the detection accuracy, resulting in increase of true positives and a notable reduction of false positives. This is illustrated in~\cref{fig:ablation_study} (second row), where a small buoy in the distance is detected only by the network variants that use the separation loss during training (WaSR$_{\mathrm{NOIMU}}$ and WaSR). The separation loss also improves segmentation of near-by large objects, which is illustrated on an example of a pier in~\cref{fig:ablation_study} (third row). Benefits are also apparent in water-edge estimation accuracy when the USV faces towards mainland or large proximal obstacles.
 

Improvements of water-edge estimation from IMU fusion are most apparent when the USV faces the open water. An example in~\cref{fig:ablation_study} (first row) shows that the water edge is strongly overestimated when not using the IMU, leading to miss-classifying an entire island on the far-left side. Similarly, in~\cref{fig:ablation_study} (second row) the water edge above the dinghy is more accurately estimated when the IMU is used.  
  
A failure case is illustrated in~\cref{fig:ablation_study} (last row). All variants of WaSR experience segmentation difficulties. Even though the IMU fusion improves the estimated water edge, part of it is still under-estimated and a small false-positive is detected on the wake. While this type of miss-classification does not lead to USV collision it clearly shows room for further improvements.


\vfill\null
\section{Conclusion} \label{sec:conclusion}


A novel obstacle detection deep neural network, WaSR, for USV navigation was presented. WaSR improves the water-edge segmentation and overall obstacle detection by fusing visual information with inertial sensory data from an on-board IMU. A deep encoder extracts rich visual features from the input image, while a non-symmetric and shallow decoder fuses the visual features with inertial data. Additional robustness is achieved by introducing a novel water-obstacle separation loss at the end of the encoder, which enforces
learning a feature space in which separation between water and obstacle appearances is increased.

Experimental results show that WaSR outperforms the state-of-the-art by over 14\% in F-measure. Compared to the second-best method  BiSeNet~\cite{yu2018bisenet}, WaSR increases true positives by 8\%, and reduces false-positives and false-negatives by 64\% and 69\%, respectively. Water edge estimation accuracy is increased by three pixels, which means that the obstacle localization error is reduced by several hundred of meters for the obstacles close to horizon. Ablation study further validated the importance of individual design choices of WaSR, in particular, the new water-obstacle segmentation loss and IMU fusion pipeline.

Our future work will focus on further speeding up the segmentation, while maintaining the accuracy. Given a significant performance boost on the USV domain, it will be interesting to test whether the architecture generalises to other, non-USV, maritime~\cite{smd_prasad2017video,ipatch_dataset} and AGV~\cite{cityscapes_dataset} scenarios.

\clearpage

{\small
\bibliographystyle{IEEEtran}
\bibliography{bibliography}
}

\end{document}

%% file: MODD2_detections.tex

\small{
\resizebox{\ifdim\width>\columnwidth\columnwidth\else\width\fi}{!}{%
\begin{tabular}{lccccc}
Architecture                               & $\mu_{\mathrm{edg}}$ & TP       & FP        & FN       & F-measure \\
\midrule
PSPNet~\cite{zhao2017pyramid}                & 13.8 (16.0)          & 5886     & 4359      &  431     &  71.1 \\ 
SegNet~\cite{badrinarayanan2017segnet}     & 13.5 (18.5)          & 5834     & 2139      &  483     &  81.7 \\ 
DL2$_\mathrm{NOCRF}$~\cite{chen2016deeplab}       & 12.8 (21.4)          & 3946     &   \textbf{227}      & 2371     &  75.2 \\
DL3+~\cite{chen2018encoder}           & 14.1 (20.9)          & 5311     & 2935      &  1006    &  72.9 \\
BiSeNet~\cite{yu2018bisenet}               & 12.4 (19.2)          & 5699     & 1894      &  618     &  81.9 \\ 
WaSR                                     &  \textbf{9.6} (18.5)          & \textbf{6166}     &  679      &  \textbf{151}     &  \textbf{93.7} \\ 
\end{tabular}
}%
}

%% file: MODD2_ablation.tex

\small{
\resizebox{\ifdim\width>\columnwidth\columnwidth\else\width\fi}{!}{%
\begin{tabular}{lccccc}
Architecture                               & $\mu_{\mathrm{edg}}$ & TP       & FP        & FN       & F-measure   \\
\midrule
WaSR                                       &  \textbf{9.6} (18.5)        & \textbf{6166}     & 679      &  \textbf{151}     &  93.7 \\ 
WaSR$_{\mathrm{NOWS}}$ & 12.3 (18.0)        & 4149     & 710      &  2168    &  74.2 \\
WaSR$_{\mathrm{NOIMU}}$     & 11.2 (17.7)        & 5943     & \textbf{296}      &  374     &  \textbf{94.7} \\
\end{tabular}
}%
}